\documentclass{article}

\PassOptionsToPackage{numbers, compress}{natbib}
\usepackage[final]{nips_2016}

\usepackage[utf8]{inputenc} 
\usepackage[T1]{fontenc}    
\usepackage{hyperref}       
\usepackage{url}            
\usepackage{booktabs}       
\usepackage{amsfonts}       
\usepackage{nicefrac}       
\usepackage{microtype}      

\def\etal{\emph{et al.~}}
\usepackage{graphicx}
\usepackage{epsfig}
\usepackage{epstopdf}

\usepackage[normalem]{ulem}
\usepackage[dvipsnames]{xcolor}

\title{Identifying and Categorizing Anomalies in Retinal Imaging Data}

\author{
   Philipp~Seeb{\"o}ck, Sebastian~Waldstein, Sophie~Klimscha, Bianca~S.~Gerendas,\\
   \textbf{ Ren{\'e}~Donner, Thomas~Schlegl, Ursula~Schmidt-Erfurth and Georg~Langs} \\
   Christian Doppler Laboratory for Ophthalmic Image Analysis, Department for Ophthalmology\\
   CIR - Department of Biomedical Imaging and Image-guided Therapy, Medical University of Vienna\\
   \texttt{philipp.seeboeck@meduniwien.ac.at} \\
}

\begin{document}

\maketitle

\vspace{-0.6cm}
\section{Introduction}
\vspace{-0.1cm}
The detection of diagnostically relevant markers in imaging data is critical in medicine and treatment guidance. Typically, detectors are trained based on a priori defined categories, and annotated imaging data. This makes large-scale annotation necessary which is often not feasible, limits the use to known marker categories, and overall slows the process of discovering novel markers based on available evidence. Additionally, in contrast to natural images such as photographs, the relevant categories of retinal images cover only a small fraction of the overall imaging data, though they are the focus of diagnostic attention, and are often dominated by natural variability of even healthy anatomy.
 
Optical Coherence Tomography (OCT) \cite{huang1991optical} is an important diagnostic modality in ophthalmology and offers a high-resolution 3D image of the layers and entities in the retina. Each position of the retina sampled by an optical beam results in a vector, the A-scan. Adjacent A-scans form a B-scan, which in turn form the entire volume. Anomaly detection \cite{pimentel2014review} in retinal images is a difficult and unsolved task, though many patients are affected by retinal diseases that cause vision loss (e.g. age-related macular degeneration has a prevalence of 9\% \cite{wong2014global}). We propose to identify abnormal regions which can serve as potential biomarkers in retinal spectral-domain OCT (SD-OCT) images, using a Deep Convolutional Autoencoder (\emph{DCAE}) trained unsupervised on healthy examples as feature extractor and a \emph{One-Class SVM} to estimate the distribution of normal appearance. By using only healthy examples for training, we omit the need of collecting a dataset that represents all the variabilities which may occur in the data and contains sufficient amount of anomalies.

Results show that the trained model is able to achieve a dice of $0.55$ regarding annotated pathologies in OCT images. Since there are different pathologies and structures occurring in the regions detected as anomalous, a meaningful sub-classification of these areas is of particular medical interest. Therefore we further cluster regions identified as anomalous in order to retrieve a meaningful segmentation of these areas. In addition, we evaluate to which extent the learned features can be used in a classification task, where intraretinal cystoid fluid (IRC), subretinal fluid (SRF) and the remaining part of the retina are classified. A classification accuracy of 86.6\% indicates the discriminative power of the learned feature representation.

\vspace{-0.1cm}
\paragraph{Related Work}
Deep Convolutional Neural Networks (CNNs) in combination with large quantities of labeled data have recently improved the state-of-the-art in various tasks such as image classification \cite{szegedy2015going} or object detection \cite{ren2015faster}. CNNs can automatically learn translation invariant visual input representations, enabling the adaptation of visual feature extractors to data, instead of manual feature engineering. While purely supervised training of networks is suitable if labeled data is abundant, unsupervised methods enable the exploitation of unlabeled data \cite{cho2015unsupervised,doersch2015unsupervised,dosovitskiy2014discriminative,zhao2015stacked}, discovering the underlying structure of data.

Doersch \etal~\cite{doersch2015unsupervised} use spatial context as supervisory signal to train a CNN without labels, learning visual similarity across natural images, which does not necessarily extend to medical domain. 
Dosovitskiy \etal~\cite{dosovitskiy2014discriminative} train a CNN unsupervised by learning to discriminate between surrogate image classes which are created by data augmentation. A limitation of this approach is that it does not scale to arbitrarily large amounts of unlabeled data. Zhao \etal~\cite{zhao2015stacked} propose joint instead of layer-wise training of convolutional autoencoders, but they perform experiments in a semi-supervised setting.

A variety of anomaly detection techniques are reported in the literature. Carrera \etal~\cite{carrera2015detecting} use the technique of convolutional sparse models to detect anomalous structures in texture images. In the work of Erfani \etal~\cite{erfani2016high} Deep Belief Networks are combined with One-Class SVMs in order to address the problem of anomaly detection in real-life datasets. In contrast to our paper, these works address problems regarding natural images and real-life datasets, which have considerable different characteristics compared to medical images, as explained above. An overview for anomaly detection methods can be found in \cite{pimentel2014review}.

Regarding unsupervised learning in OCT images, Venhuizen \etal~\cite{venhuizen2015automated} train random forests with features from a bag-of-words approach in order to classify whole OCT volumes, as opposed to our pixel-wise segmentation approach.
Schlegl \etal~\cite{schlegl2015automatic} use weakly supervised learning of CNNs to link image information to semantic descriptions of image content. In contrast to our work, the aim is to identify a priori defined categories in OCTs using clinical reports.

\section{Method}
\vspace{-0.13cm}
The following preprocessing is applied to all volumes. First we identify the top and bottom layer of the retina using a graph-based surface segmentation algorithm \cite{garvin2009automated}, where the bottom layer is used to project the retina on a horizontal plane. Then each B-scan is brightness and contrast normalized. Finally we perform over-segmentation of all B-scans to monoSLIC superpixels $sp$ of an average size of $4\times4$ pixels \cite{mholzer2014superpixel}.

To capture visual information at different levels of detail, we use a multi-scale approach to perform superpixel-wise segmentation of the visual input. We conduct unsupervised training of two DCAEs on pairs of extracted patches sampled at the same positions for two scales ($32\times32$ patches for $ DCAE_{1} $ and down-sampled $128\times32$ to $32\times32$  patches for $ DCAE_{2} $) in parallel.
The used network architecture for both scales is \texttt{512c9-3p-2048f-512f} for the encoder, implying a matching decoder structure that uses deconvolution and unpooling operations. All layers except pooling and unpooling are followed by Exponential Linear Units (ELUs) \cite{clevert2015fast}. The loss function for training is defined as the Mean Squared Error function $ MSE(x, \hat{x}) $, where $ x $ denotes the input patch and $ \hat{x} $ the reconstructed input. In addition, we use dropout in each layer, which corresponds to unsupervised joint training with local constraints in each layer.

The feature representations of both scales are concatenated and used as input for training a Denoising Autoencoder ($ DAE $), its single-layer architecture denoted as \texttt{256f}. All three models together form our final model $ DCAE_{ent} $ that gives us a 256 dimensional feature representation $ z $ for a specific superpixel in the B-Scan. 

We then train the One-Class SVM \cite{scholkopf2001estimating} using a linear kernel to find a boundary that describes the distribution of healthy examples in the feature space $ z $, which serves as decision boundary for unseen data. New samples can be classified either as coming from the same data distribution if lying within the boundary (normal) or not (anomaly). For each OCT, features $z$ and the corresponding class are computed for each superpixel lying within the top and bottom layer of the retina. This provides a segmentation of the retina into two classes.
 
Subsequently, we use spherical K-means clustering~\cite{hornik2012spherical} with cosine distance to sub-segment regions which have been identified as anomalous in the former step into multiple clusters $c$.
The number of cluster centroids is determined by an internal evaluation criterion called Davies-Bouldin (DB) index~\cite{halkidi2001clustering}, where a small value indicates compact and well separated clusters.
To segment an unseen OCT, each superpixel with the property ''anomalous'' gets a cluster assignment, according to the nearest cluster centroid in the feature space $z$.

\section{Evaluation}
\vspace{-0.13cm}
The primary purpose of the evaluation is to test if we can identify novel marker candidates in imaging data algorithmically, instead of relying on a priori defined object categories. We evaluated (1) if we can segment anomalies, (2) if we can find categories of these regions that correspond to a fine-grained ontology of known findings, and (3) if the learned features have discriminative power for image classification.

We used scans from 704 patients, where 283 healthy OCT volumes were used to create the healthy training set (277,340 pairs of image patches) for the $DCAE$ and the SVM, 411 unlabeled OCTs were used to create the anomaly training set (295,920 patches) for clustering, and the validation and test set consisting of 5 volumes each, with voxel-wise ground truth annotations of anomalous regions and additional annotations for specific pathologies (IRC, SRF), were used for model selection and evaluation, respectively.
The scans were acquired using Spectralis OCT instruments (Heidelberg Engineering, GER) and have a resolution of $512\times496\times49$ depicting a $6mm\times2mm\times6mm$ volume of the retina, where the distance between voxel centers is about $11{\mu}m$ in the first, $4{\mu}m$ in the second, and $120{\mu}m$ in the third dimension.

The learned anomaly detection model is evaluated on the test set, where \emph{Dice}, \emph{Precision} and \emph{Recall} are calculated for anomalous regions regarding the ground-truth annotation. Interpretation of these quantitative values is done carefully and only in combination with qualitative evaluation by clinical retina experts.

In addition, we compare our $ DCAE_{ent} $ model with conventional PCA embedding. To ensure fair comparison, we train two models. In the first model $PCA_{256}$, the dimensionality is chosen to match the feature dimension $z$ of our proposed model. For both scales, the first 128 principal components are kept and concatenated to obtain $ z $. In the second model $PCA_{0.95}$, for each scale the first components that describe 95\% of the variance in the dataset are kept.

For the categorization of anomalous regions the number of clusters is varied between 2 and 30, where the model with the lowest DB-index on the anomaly training set is selected and applied to the test set.
We qualitatively evaluate if the categories found in the regions identified as abnormal are clinically meaningful, to provide a link to actual disease.

Beside anomaly detection, we test the learned models in a classification task in order to evaluate the discriminative power of the learned feature representation. Here, we train a linear Support Vector Machine (L2-SVM) on a balanced classification dataset with 15,000 labeled examples and three classes (33.3\% IRC, 33.3\% SRF, 33.3\% remaining part of the retina) in feature space $z$. 5-fold-cross-validation is performed so that samples of a patient are only present in one fold, where this labeled dataset is only used to train the L2-SVM.

\section{Results}
\vspace{-0.13cm}
We report quantitative and qualitative results illustrating anomaly segmentation, visualize anomaly categorization outcome, provide descriptions of clusters according to the experts and describe results regarding the classification task.

\begin{table}[tt]
	\parbox{.42\linewidth}{
	\caption{Dice, Precision and Recall for anomalous regions with ground-truth annotations. The same One-Class SVM settings are used for all methods.}
	\label{tab:result_anomaly}
	\begin{tabular}{llll}
		\toprule
		Algorithm & Dice & Precision & Recall \\
		\midrule
		$PCA_{256}$ &  0,33 & 0,31 & 0,39 \\
		$PCA_{0.95}$  &   0,32 & 0,32 &  0,35 \\
		$DCAE_{ent} $ &  0,55 & 0,53  & 0,58 \\
		\bottomrule
	\end{tabular}
}
\hfill
\parbox{.53\linewidth}{
	\caption{Mean classification accuracies of L2-SVMs on the generated OCT-dataset with balanced classes, trained with features from different models.}
	\label{tab:result_classification}
	\begin{tabular}{lllll}
		\toprule
		& \multicolumn{4}{c}{Accuracy (in percent)} \\
		\cmidrule{2-5}
		Algorithm & IRC & SRF & Other & Overall\\
		\midrule
		$PCA_{256}$ & 71.9 & 74.0 & 73.9 &  73.4 ($\pm$ 3.9)   \\
		$PCA_{0.95}$  & 73.3 & 76.9 & 70.0 & 73.4 ($\pm$ 3.9)    \\
		$DCAE_{ent}$ & \textbf{87.3 }& \textbf{88.3} & \textbf{84.1} & \textbf{86.6 ($\pm$ 1.6)}\\
		\bottomrule
	\end{tabular}
}
\vspace{-0.3cm}
\end{table}

\begin{figure}[t]
	\centering
	\includegraphics[height=7.5cm,clip=true,trim=5.5cm 7.2cm 7.5cm 4cm]{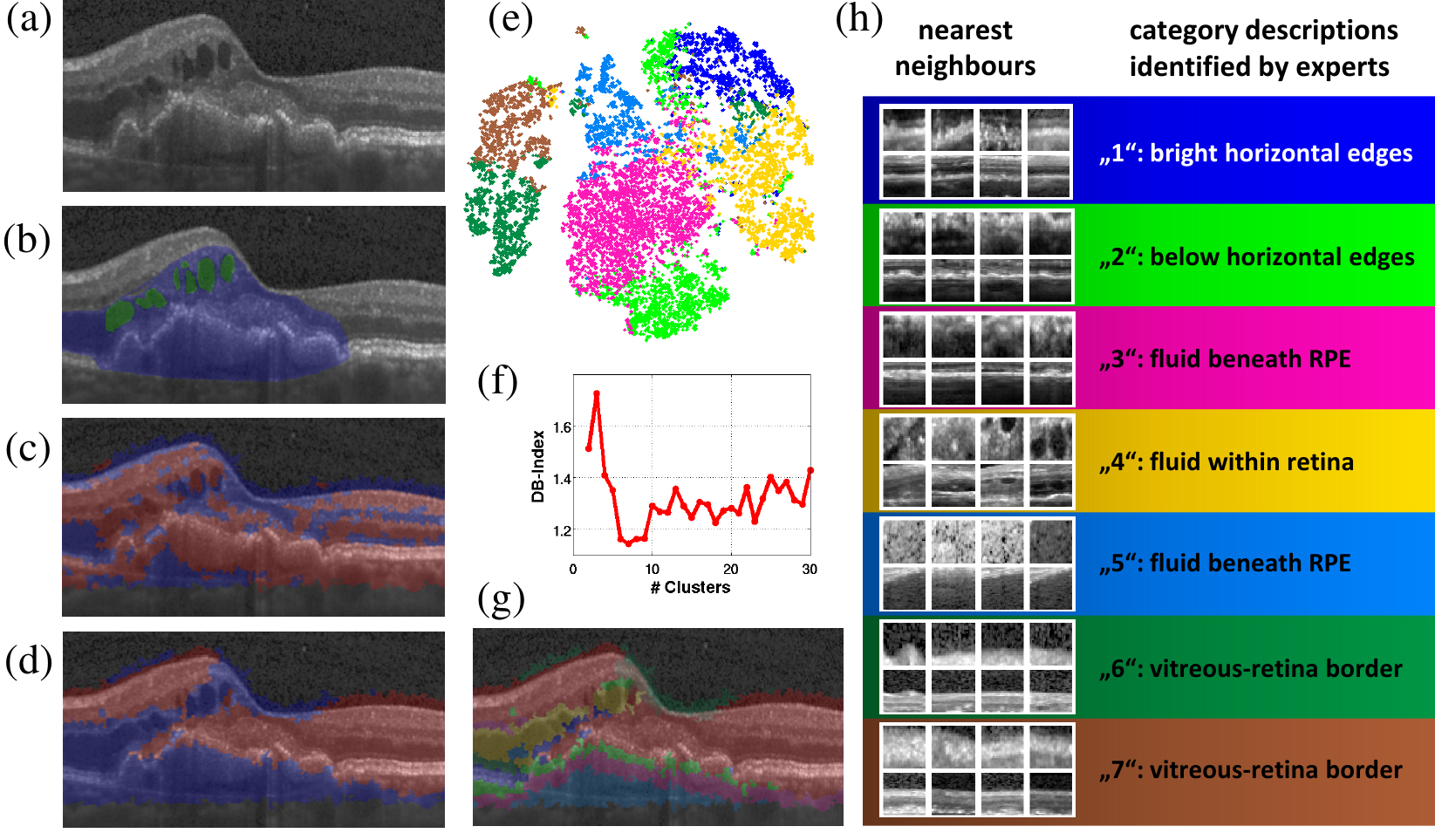}
	\caption{The same B-scan is illustrated for (a) the original scan, (b) the ground truth annotations (pathologic region in blue, IRC in green), anomaly detection results for (c) the comparison method $PCA_{0.95}$ and (d) our proposed method (normal=red, anomaly=blue). The clustering result is shown in (g), where identified anomalous regions are segmented into 7 categories. Each cluster is indicated by a separate color. The corresponding cluster descriptions which are identified by experts, are illustrated in (h) together with nearest neighbors of cluster centroids. 2D t-SNE embedding of the feature space $z$ is shown in (e). The calculated values of the DB-Index are plotted in (f). }
	\label{fig:result_clusteringOCT}
	\vspace{-0.3cm}
\end{figure}

As can be seen in Table \ref{tab:result_anomaly} our method achieves a dice of $0.55$, which is a clear improvement in comparison with both $PCA_{256}$ and $PCA_{0.95}$. This is also reflected by the visualization provided in Figure \ref{fig:result_clusteringOCT} (a)-(d), which shows substantial overlap between ground truth annotations and  $DCAE_{ent}$ segmentation. $DCAE_{ent}$ provides a less diffuse segmentation compared to $PCA_{0.95}$, capturing the retinal pathology in a meaningful way.

Regarding anomaly categorization, the lowest DB-index was found for 7 clusters, as illustrated in Figure \ref{fig:result_clusteringOCT} (f). Figure \ref{fig:result_clusteringOCT} (e) shows 2D t-SNE embedding \cite{van2008visualizing} of the learned $ DCAE_{ent} $ features.
Categories were identified and described as following by two clinical retina experts and are summarized in Figure \ref{fig:result_clusteringOCT} (g)-(h). 

Bright horizontal edges are segmented in cluster ''1'' which is highlighted in blue. In the majority of cases this cluster corresponds to Retinal Pigment Epithelium (RPE). Cluster ''2'' is marked in light green and corresponds to areas below these horizontal structures. Both clusters segment areas situated next to fluid, which changes the local appearance of patches to abnormal.
Cluster ''4'' is highlighted in yellow and corresponds to fluid within the retina. This finding is supported by the fact that 62\% of manual IRC annotations located in anomalous regions are assigned to cluster ''4''.
Marked in grey-blue and pink, Cluster ''3'' and ''5'' both segment regions that correspond to fluid beneath the RPE.
Cluster ''6'' (dark green) and ''7'' (brown) highlight the border between vitreous and retina due to irregular curvature or fluid situated below, which alters the appearance of extracted patches.

The classification results are illustrated in Table~\ref{tab:result_classification}, where mean overall accuracy as well as class specific accuracies are reported.
The proposed unsupervised feature learning approach $ DCAE_{ent} $  clearly outperforms both conventional methods $PCA_{256}$ and $PCA_{0.95}$. Our proposed method achieves a mean overall accuracy of 86.6\%, while the PCA-models only achieve 73.4\% and 73.3\%.

\section{Discussion}
\vspace{-0.1cm}
In this paper we propose a method to detect anomalous regions in OCT images which needs only healthy training data. A deep convolutional autoencoder is used as feature extractor, while One-Class SVM is used to estimate the distribution of healthy retinal patches. In a second step, the identified anomalous regions are segmented into subclasses by clustering. Results show that our proposed method is not only capable of finding anomalies in unseen images, but also categorizes these findings into clinically meaningful classes. The identification of new anomalies, instead of the automation of expert annotation of known anomalies is a critical shift in medical image analysis. It will impact the categorization of diseases, the monitoring of treatment and the discovery of relationships between genetic factors and phenotypes. Additionally, the power of our learned feature extractor is also indicated by performance in a classification task.

\vspace{-0.15cm}
\subsubsection*{Acknowledgments}
\vspace{-0.2cm}
This work funded by the Austrian Federal Ministry of Science, Research and Economy, and the FWF (I2714-B31). A Tesla K40 used for this research was donated by the NVIDIA Corporation.


\medskip

\small

\bibliographystyle{abbrv}
\bibliography{nips_2016_references}

\end{document}